\documentclass[11pt]{article}

\usepackage[preprint]{acl}

\usepackage{times}
\usepackage{latexsym}

\usepackage[T1]{fontenc}

\usepackage[utf8]{inputenc}

\usepackage{microtype}

\usepackage{inconsolata}

\usepackage{graphicx}

\title{We can still parse using syntactic rules }

\author{Hussein Ghaly \\
  Independent Researcher  \\
  New York, USA \\
    \texttt{hmghaly@gmail.com} \\}

\begin{document}

\maketitle
\begin{abstract}
This research introduces a new parsing approach, based on earlier syntactic work on context free grammar (CFG) and generalized phrase structure grammar (GPSG). The approach comprises both a new parsing algorithm  and a set of syntactic rules and features that overcome the limitations of CFG. It also generates both dependency and constituency parse trees, while accommodating noise and incomplete  parses. The system was tested on data from Universal Dependencies, showing a promising average Unlabeled Attachment Score (UAS) of 54.5\% in the development dataset (7 corpora) and 53.8\% in the test set (12 corpora). The system also provides multiple parse hypotheses, allowing further reranking to improve parsing accuracy. This approach also leverages much of the theoretical syntactic work since the 1950s to be used within a computational context. The application of this approach provides a transparent and interpretable NLP model to process language input.  
\end{abstract}

\section{Introduction}

With all the advances in Transformers and LLMs, do we still need syntactic parsing? 

In recent years, most advanced NLP applications, such as Automatic Question Answering and Machine Translation seem to have been implemented without a need for explicit syntactic parsing. In fact, transformers with their attention mechanism \cite{vaswani_attention_2017} can implicitly infer the structure and dependencies of words and tokens in any language input.

However, while transformer-based models typically achieve strong performance on various NLP tasks, their syntactic decisions are not explicit. As a result, it is often unclear why a particular structure is produced or how certain syntactic rules apply across constructions, domains, or languages. 

Furthermore, with their massive computational requirements and difficult explainability, there is rationale to re-examine  parsing as an essential NLP task to reveal the structure of human language. This task has traditionally been quite challenging in several ways, both computationally and conceptually. Perhaps one of the biggest challenges is  the lack of a universal theoretical syntactic framework to guide the parsing process. The following is a brief review of the syntactic theory involved.

In 1956, Noam Chomsky published his paper, "Three Models for the Description of Language" \cite{chomsky_three_1956}, and soon after his landmark book “syntactic structures” \cite{chomsky_syntactic_1957}. His main motivation for this work was: 

“We investigate several conceptions of linguistic structure to determine whether or not they can provide simple and "revealing" grammars that generate all of the sentences in English and only these”

This entailed a quest to find all the rules that could be used to generate or process any sentence in English, allowing the identification of  any grammatical and ungrammatical sentence as such.

The second model investigated by Chomsky was Context Free Grammar (also referred to as Phrase Structure Grammar), where the grammar can be described in terms of rewrite rules, such as NP → Det Adj Noun (a noun phrase is made of a determiner, adjective and noun). However, he considered this model inadequate for representing human language because it does not account for phenomena such as syntactic gaps.

Instead, he proposed a model based on transformation and movement of syntactic elements, commonly referred to as Transformational Grammar (TG), Generative Grammar (GG) or Universal Grammar (UG). The main premise is that there is a deep structure of the sentence, and there is a surface structure of how it is produced after applying certain transformations. These transformations are context sensitive, precluding using direct CFG rules for representing human sentences. Research in this generative field underwent different phases and simplifications, culminating into Chomsky’s “the Minimalist Program”  \cite{chomsky_minimalist_1995}, which simplified the transformations and sentence representations, but still kept the main paradigm of transformation.

One of the main contrasting ideas to Transformational Grammar was Generalized Phrase Structure Grammar (GPSG) \cite{gazdar_generalized_1985}, and the subsequent Head-driven phrase structure grammar (HPSG) \cite{pollard_head-driven_1994}. The main assertion of GPSG was that it is possible to represent human sentences and grammar using certain syntactic rules and features, without the need to invoke any kind of transformation. This idea was demonstrated in addressing the syntactic gap phenomenon, by positing what is known as “slash features”, rather than any kind of movement.

Example: The man I met X

In this example, the word “met” is a verb phrase missing a noun phrase object, so it is represented as VP/NP, and it is projected to “I met”, which is a sentence missing a noun phrase S/NP. Hence it is possible to represent this example above using a rule such as NP → NP S/NP.

This approach could have helped re-introduce the possibility of parsing and representing the language based on syntactic rules. However, it was shown that movements and transformations are indeed sometimes necessary, as demonstrated in the cases of Dutch and Swiss German \cite{shieber_evidence_1985}.

With modern parsing after 2000, limited use of syntactic rules has been observed. Instead, most of the focus was on data-driven approaches without reliance on explicit rules. 
\begin{figure}
    \centering
    \includegraphics[width=1\linewidth]{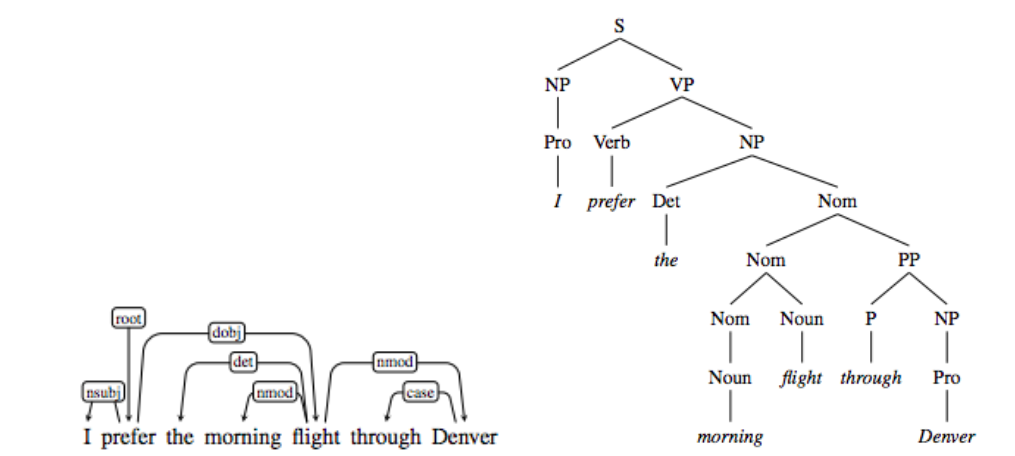}
    \caption{Dependency vs. Constituency Structures  
    (From \cite{martin_vector_2019} }
    \label{fig:dep-const-fig}
\end{figure}

Another important consideration for modern parsing is the fact that two different paradigms are used: Constituency Parsing and Dependency Parsing. As we can see in figure 1, constituency parsing mainly focused on identifying  hierarchical constituents/phrases as the underlying structure of a sentence (e.g. Noun Phrase, Verb Phrase .. etc). Dependency parsing provides a representation of the dependency relations between words of the sentence. Most of the theoretical work in syntax used consituency structures, while there was a strong focus on dependency structures in modern parsing applications. 

Each paradigm has a different focus and can provide different information. This combined information can be useful for a variety of applications. It has been shown that it is possible to convert from one paradigm to the other and vice versa \cite{xia_converting_2001}. However, constituency structure does not necessarily include information about the head of each phrase, making it more difficult to convert from constituency to dependency. Also, dependency structure does not have a clear method of projecting phrases from words. Therefore, it is important for any parser to capture both representations, and this should be built into the parsing algorithm itself. 

In addition, there are several practical challenges for parsing sentences in realistic contexts.  For example, spoken sentences usually include disfluencies, errors and speech repairs (e.g. ummm, you know), and similar noise, so a parser needs to accommodate these elements. Also, in written text, there are extra-textual elements, such as different kinds of punctuation marks, in addition to elements such as footnotes, hyperlinks, formatting tags (e.g. bold <b>, italic <i>), and so forth.  Furthermore, elements such as emojis and HTML entities can be challenging to parse. All these elements enrich the information encoded in text, so a parser needs to be able to handle them.

To summarize, based on the challenges outlined above, the main motivation of this research is to develop a theoretically valid syntactic parsing approach that can process different linguistic structures, while dealing with real-world text input, with the goal to produce a dual-paradigm structure of language encompassing both dependency and constituency structures. 

\section{Related Work}

The current goal of this research is to build on theoretical linguistic formalisms for automatic syntactic parsing. Over the past decades, we find a number of similar approaches, building on GPSG/HPSG formalisms, as well as those by Chomsky’s Minimalist Grammar  and related Transformational grammar. 

Much of this related work focused on developing HPSG parsers, such as the following recent attempts:

\begin{itemize}
    \item  \citealt{nguyen_attempt_2024}: developed a Neural Parser based on Simplified HPSG rules on Vietnamese, with mixed results. HPSG
    \item \citealt{zamaraeva_revisiting_2024}: developed a BERT-based supertagger using HPSG rules for constituency parsing
    \item \citealt{li_head-driven_2021}: trained HSPG parser from only constituency and dependency annotations
    \item \citealt{torisawa_hpsg_2000}: introduced an HPSG parser using a technique called CFG filtering to reduce computational parsing costs
    \item \citealt{zhou_head-driven_2019}: attempted to formulate a simplified HPSG by integrating constituent and dependency formal representations into head-driven phrase structure
    \item \citealt{yoshida_corpus-oriented_2005}: reported corpus oriented development of a wide coverage Japanese HPSG parser
    \item \citealt{fisher_practical_1989}: developed a parsing algorithm for GPSG 
\end{itemize}

In addition, certain approaches focused on Minimalistic and Transformational Grammar, such as the following:

\begin{itemize}
    \item \citealt{chesi_expectation-based_2021}: Developed a framework for formalizing Minimalist Grammar from a parsing perspective
    \item \citealt{berwick_minimalist_2019}: this book introduced demonstrations on how to turn Chomsky’s abstract theories into working computer programs
    \item \citealt{torr_wide-coverage_2019}: presented an application of Minimalist Grammar formalism to the task of realistic wide-coverage parsing
    \item \citealt{indurkhya_automatic_2019}: introduced a parser for Minimalist Grammar and an approach for inferring Minimalist Grammar rules
    \item \citealt{portelance_grammar_2017}:  presented a formal approach to probabilistic minimalist grammar parameter estimation, and an algorithm for the application of variational Bayesian inference to this formalization of Minimalistic grammar.
    \item \citealt{fowlie_parsing_2017}: presented an approach of parsing with Minimalist Grammar
    \item \citealt{mainguy_probabilistic_2010}: proposed a probabilistic top-down parser for minimalist grammars, and an approach for rewriting Minimalist Grammars as Linear Context-Free Rewriting Systems
    \item \citealt{harkema_recognizer_2000}: presented a bottom up parsing method for Minimalist Grammars
\end{itemize}

These approaches mainly introduce the theoretical aspects of parsing. They include different implementations and discussions of the computational complexity involved. However, there are only very limited attempts to apply these approaches to real world data such as corpora and treebanks, and report the results against other state-of-the-art data-driven parsing approaches. 

\section{Method}

The goal in this line of research is to develop a parsing method that takes any language input and produces a sentence structure that is interpretable and theoretically valid, while addressing several practical and theoretical challenges :

\begin{itemize}
    \item Parse the sentence according to a set of predefined syntactic rules
    \item Produce a structure that captures both dependency and constituency information
    \item Address noise , such as speech disfluencies (e.g. The book, uh, that I read), and unparseable elements, as well as punctuation marks and other extra-textual elements
    \item Address theoretical structural challenges, such as syntactic gaps (e.g. The man that I saw x)
    \item Address ambiguity and the possibility that there can be more than one underlying valid structure
\end{itemize}
This parsing method is evaluated qualitatively and quantitatively. Qualitative evaluation is performed by inspecting the output of the parser and checking whether it reflects the different predefined structural rules or not. Quantitative evaluation involves testing the parser on multiple corpora and measuring the actual parsing accuracy, and reporting the performance against other systems. 

\subsection{Parsing rules}

This research proposes a number of new rules that can be used to parse English sentences. Such rules are compatible with the considerations of GPSG. For example:

Parent\_Category[feature1] → Child1\textasciicircum[feature] Child2[feature] 

Each rule has one parent category, and either one or two child categories. Each category can have a number of positive or negative features. The notation (\textasciicircum)     is used to indicate the head child of the current rule.

These rules are flexible, with custom categories and features that allow avoiding the shortcomings of CFG.

For example, noun phrases can be compounded, for things such as (United Nations Development Program), so it is possible to propose a rule such as:

NP → NP NP

However, this can allow phrases such as (my brother the book *), so an important consideration is to make sure that the noun phrase (NP) does not include a determiner, so in this case, we propose a new category NP-U, for an NP without a determiner.  In this case, we can change the rule above to the following:

NP → NP-U NP-U 

Additionally, we can use these categories for the generation of noun phrases with an adjectival phrase (AP):

NP → AP NP-U

Also, noun phrases with determiners (DT) need these categories:

NP-DT → DT NP-U

Then eventually both NP-DT and NP-U can be projected to a normal NP:

NP → NP-DT

NP → NP-U

A similar situation applies for verbs and verb phrases, because it is possible to assume a CFG rule as follows:

VP → V (a verb phrase is projected from a verb)

VP → VP NP (a verb phrase is projected from a verb phrase followed by a noun phrase)

However, this last rule can allow successive recursion and allows a verb phrase to contain more than two noun phrase objects, which does not happen in English. Therefore, the following modified rule is created:

VP-O → V NP (projecting to a category of a VP with one object)

VP-2O → VP-O NP (projecting to a category of a VP with two objects)

Then, both transitive/ditransitive verb phrase categories (VP-O and VP-2O), as well as non-transitive verb phrase (V), can all be projected to VP

For auxiliaries, it is also necessary to identify the set of rules and categories that can allow the correct sequences of English auxiliaries (e.g. would have been) and disallow incorrect sequences (e.g. was is can *).

For the verb “to be”, it can be either a stand alone (is, was, were .. etc), preceded by modal (e.g. will be), preceded by any form of verb “to have” (e.g. have been, had been), which can in turn be preceded by modal (will have been). There is also the form “is being”, but in this case, the verb “being” can be treated like any other “-ing” verbs. Therefore, we proceed as follows:

Any auxiliary verb with [+be] feature projects to BE category:

BE → AUX[+be]

Any combination of the verb “have” and “been” also projects to BE:

BE → HAVE AUX[+been]

The combination also works when preceded by modal (MD), the infinitive feature can be added to ensure agreement:

BE → MD BE

Adverbs can be added before or after BE categories (is really - really is) and V categories (quickly walks - walks quickly).

\subsection{Parsing Algorithm}

The proposed parsing algorithm follows an incremental parsing approach for a sequence of tokens and their likely Parts of Speech tags. Each tag or category is projected to a higher category using a set of syntactic rules.

\subsubsection{Tokenization}

The current parser accepts the inputs as tokens from any tokenization system. One of its main considerations is to be flexible about the variations in tokenization outputs; for example, for contractions, hyphenation, and numbers. 

\subsubsection{POS Tagging}

For each token, the system identifies its likely POS tags, along with the probabilistic weight of each of these tags. This is an important feature of the current parser, allowing it to generate multiple parse hypotheses, along with determining the weight of each.

These tags include mainly XPOS \footnote{https://cs.nyu.edu/\~grishman/jet/guide/PennPOS.html} tags found in Universal Dependencies treebanks. It is important to use these rather than the more generic UPOS (e.g. NOUN, VERB … etc), because they are too general. For example, certain rules apply only to certain types of verbs (e.g. VBG gerund verbs have different rules from VB base form verbs). 

The system also includes hand crafted lists of tags and features for closed class words/function words (e.g. pronouns, prepositions, determiners … etc). These custom tags also allow for specific rules, such as those related in particular to the verb “to be”, the verb “to have”, among others.

Although many POS taggers are available, an important consideration of this system is to generate multiple tags with their weights, which is not always available in many popular POS tagging systems. Therefore, a custom POS tagger was developed based on the training data from multiple English datasets, using Recurrent Neural Networks (RNNs).

The input to the POS tagger is a list of tokens, and the output is the tag probabilities for each token. Certain cutoff probability can be introduced to filter out tags below a certain threshold. 

\subsubsection{Phrase Creation}

To start the actual parsing process, the input is the list of tokens and their POS tags, and any additional tags from hand crafted lists.

The algorithm starts incrementally from left to right. For each word, the parser identifies its distance from the beginning of the sentence and its corresponding tags. Accordingly, for each tag, it creates a phrase object. This phrase object includes  information as in the following example:

\{'wt': 1, 'span': 1, 'start': 5, 'end': 5, 'cat': 'IN', 'feat': ['loc'], 'head\_loc': 5, 'children': [], 'level': 0\}

\subsubsection{Phrase Indexing}

An important feature of this parser is including multiple data structures for indexing phrases. The main data structure used is a multi-key dictionary which includes both the span and the category. The values are the list of phrases that correspond to this category and span. It is structured as follows 

Dictionary[end][start][category]=[phrase1, phrase2 … ]

\subsubsection{Rule Scanning}

Based on the category (cat) of the phrase object, in addition to its features (feat), it is possible to identify the rules that can apply to this phrase. It is important to notice that rules can be either binary (one parent phrase branches into two child phrases) or unary (one parent phrase has one child phrase). In either case, the parser identifies the applicable rules by scanning all rules whose last child applies to the current phrase.

\subsubsection{Adjacent Phrase Scanning}

If the applicable rule is binary, the algorithm identifies the category of the first child of the rule, and starts scanning from previously created phrases. The algorithm checks the span of the current phrase, mainly its start index. Accordingly, the algorithm identifies a set of phrases that satisfy the current rule, in terms of being located to the left of the current phrase, and have the matching category and features of the first child of the rule.

This algorithm allows skipping tokens between adjacent phrases, allowing the parser to skip noise in the input tokens, such as disfluencies in spoken sentences.

\subsubsection{Phrase Projection}

Once a rule is satisfied, whether binary or unary, the phrase is said to be projected up into the sentence structure. This projection involves the following:

\begin{itemize}
    \item Identifying the parent category based on the applicable syntactic rule
    \item Identify the children, mainly the current phrase being projected, as well as the adjacent phrase identified 
    \item Identify the weight of the newly created phrase as the sum of the weights of the children
    \item Identify the span of the new phrase, as the combined span of the two phrases
    \item Identify the level of the new phrase as one level higher than the current phrase
    \item Identify the head phrase, and the location of the head word in the new phrase
    \item Identify the features of the new phrase according to the features of the parent in the applicable rule
\end{itemize}

Once this new phrase is created, the algorithm indexes it, and then recursively projects it, while scanning applicable rules and adjacent phrases.

\subsubsection{Connecting and Reranking}

An important aspect of this parsing approach is that it applies only to the parseable chunks of the input tokens, which in some cases leaves some chunks of tokens unparsed. Therefore, we employ a variant of Dijkstra's algorithm, in order to connect the most likely parsed chunks, mainly in case there is no output parse that encompasses the full span of input tokens.

In addition, since this parsing approach produces a number of possible phrases for each span, an important element is how to sort these phrases according to their likelihood. We employ a simple scoring and reranking algorithm, which yields lower weight to the parses with more than one head or with more unparsed tokens. More advanced scoring and reranking can be used in future work.

\subsubsection{Output and Export}

With each new projected phrase, it is added to a dictionary and to the corresponding span. The main parse will be the phrase that spans the full token sequence or most of it, in addition to its recursive children. This recursive listing of phrases and their children is converted to both dependency structure and constituency list of phrases.

\section{Experimental Setup}

\subsection{Metrics}

In order to evaluate the performance of the current parsing approach, we use Unlabelled Attachment Score (UAS). This metric reveals the quality of parsing mainly in terms of showing which words depend on which.  The measurement counts how many tokens the parser identified their heads correctly, divided by the total number of tokens in the dataset.

\subsection{Data}

This parser is tested on the dev/test datasets from Universal Dependencies (UD) \footnote{\href{https://universaldependencies.org/}{https://universaldependencies.org/} } English treebanks \cite{nivre_universal_2020}. 

It includes multiple corpora with different distributions for train/dev/test sets. It includes 8 files in the train set, 7 files in the dev set, and 13 files for the test sets. Each file includes sentences annotated according to the CoNLL format for dependency parsing. This parser is applied on the tokenized sentences in each CoNLL item from the treebank. 

  

For benchmarking, we use spaCy dependency parser \footnote{\href{https://spacy.io/api/dependencyparser}{https://spacy.io/api/dependencyparser} } \cite{honnibal_joint_2014}, which is one of the most commonly used parsers. The results are based on the SpaCy standard model, by applying it to the tokenized sentences from the treebank.

\subsection{Code}

The code used for these experiments, including notebooks, python files, Excel sheets for rules and other lists, is included in a Gitlab repository \footnote{\href{https://gitlab.com/acl2575601/acl-parsing-2026}{https://gitlab.com/acl2575601/acl-parsing-2026} } for easy access and reproducibility. The readme file includes relevant instructions for using the code. 

\begin{figure*}
        \centering
    \includegraphics[width=1\linewidth]{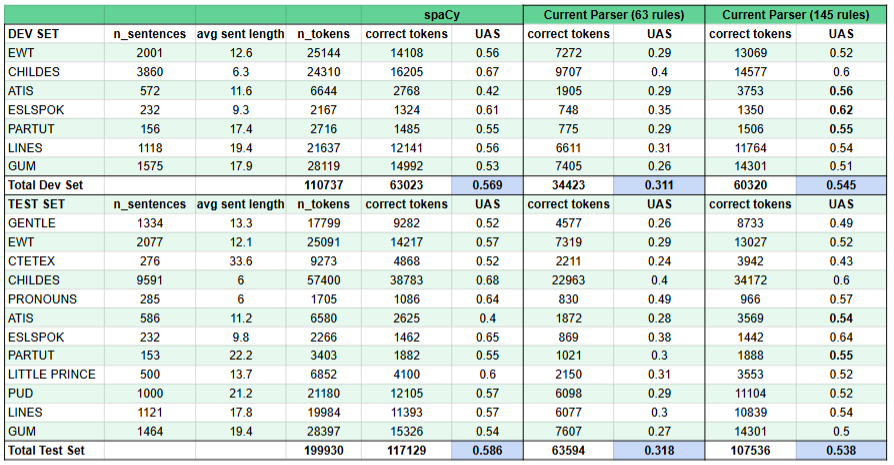}
    \caption{Parser Performance on different UD datasets, against spaCy parser performance}
    \label{fig:placeholder}
\end{figure*}
\section{Results and Discussion}

For quantitative evaluation, we compare the UAS metrics from the current parser to those from the SpaCy parser. Results are obtained on Dev/Test sections of UD datasets, and compared with the outputs from spaCy parser. As we can see from figure 2, The average UAS for current parser is 54.5\% and 53.8\% on devset and test set, respectively, compared to 56.9\% and 58.6\% by spaCy.

The performance of the current parser is promising, since we can see that it is within range of SpaCy performance. In addition, the performance is mostly robust across multiple datasets, surpassing SpaCy in multiple datasets across the devset and testset. The performance figures reflect how many phrases and dependencies are indentified correctly by the parser. 

At this stage, the parser is only programmed with a small subset of rules of English language, just as a proof of concept. These rules do not provide full coverage of English grammar or writing conventions. However, we can see that we could achieve a significant improvement by adding more rules. Using 63 rules, the performance was: devset 31.1\% - testset 31.8\%. Using 145 rules,  the performance was: devset 54.5\% - testset 53.8\% - more than 20 points improvement in each set, reflecting better coverage and accuracy.

Qualitatively, it is interesting to notice that parser output reflects both constituency and dependency structures. The representation proposed includes the tokens of the input sentence, as well as their IDs, heads and tags, which capture some of the main information from the dependency structure. It also includes a hierarchical listing of the phrases, their levels and their spans, as we can see in figure 3.
\begin{figure}
    \centering
    \includegraphics[width=1\linewidth]{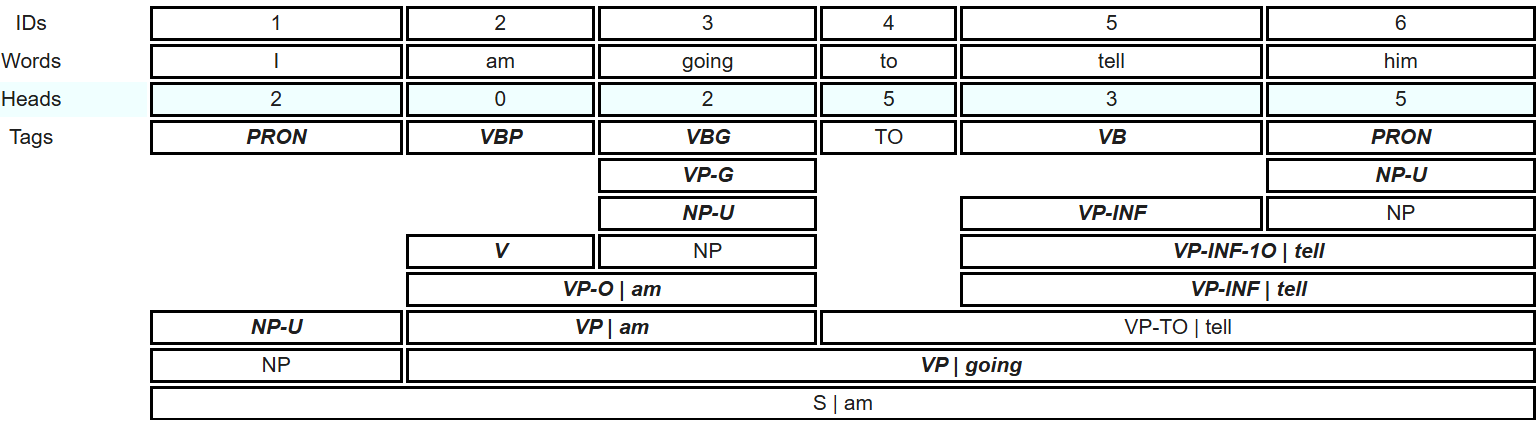}
    \caption{Illustration of parser output with the combined dependency and constituency structures }
    \label{fig:combined-format-illustration}
\end{figure}

An important point to examine is the applicability of this parsing approach to custom syntactic rules pertaining to certain linguistic structures and phenomena. For this case, we address, as an example, the slash features within GPSG. We can see in figure 4 that the parser was able to correctly handle and represent the structure of the sentence, showing all the rules applied. Such rules include the slash rule VP-O/NP corresponding to a VP missing a direct object (the gap construction), and its projection into S/NP. Eventually, we see the application of the rule:

NP → NP \textasciicircum S/NP

\begin{figure}
    \centering
    \includegraphics[width=1\linewidth]{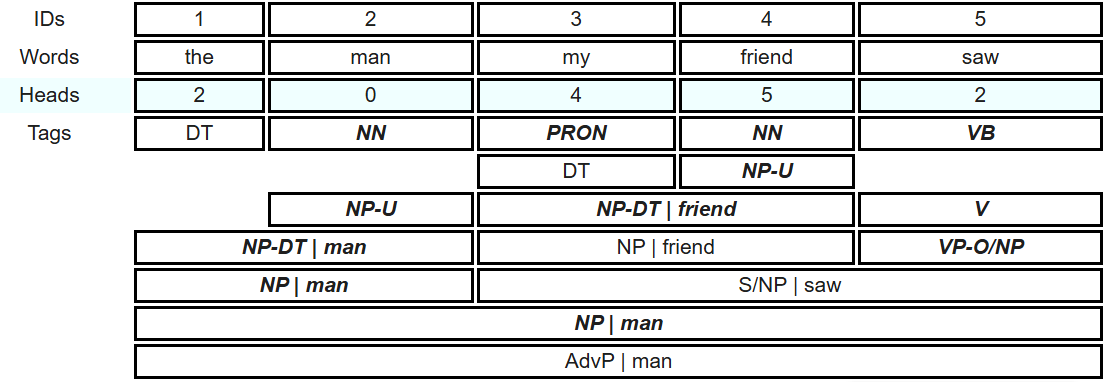}
    \caption{Illustration of parser output for applying GPSG slash features }
    \label{fig:slash-parsing-illustration}
\end{figure}

\section{Conclusions}

This current paper proposes an initial step in a new line of research. There has been a wealth of knowledge and insights over the past several decades of linguistics work, in the field of syntax and other areas of theoretical linguistics. However, most of this knowledge was not directly usable in computational applications, including parsing. Our work reconciles this gap between theoretical linguistics frameworks and computational NLP applications.

Using syntactic rules, including GPSG slash features, in this parsing approach indicates that this reconciliation is doable. In fact, this approach produces a richer combination of dependency and constituency structures, leveraging the strengths of both paradigms. It is compatible with modern parsing tools and evaluation schemes and metrics, yet more research is still needed to discover the potentials and limitations of such an approach in different kinds of data.

With this new approach, it is possible to expand to any new language. What is needed is the following:

\begin{itemize}
    \item An inventory of possible POS tags, categories and features for this language
    \item Certain CFG/GPSG rules for this language
    \item A framework for predicting POS tags for this language
    \item For languages with rich morphology, also a framework for predicting the likely subtokenization of prefixes, suffixes and subwords
\end{itemize}

Therefore, this new approach for parsing can provide a transparent, interpretable NLP model for processing language input, which can be useful for explainable AI applications and frameworks.

\section*{Limitations}

The results discussed in this paper are only related to English. The rules and features used in the system are hand crafted only as a sample and do not represent a comprehensive listing of English syntactic rules. The Parts of Speech tagger included was trained on training data from Universal Dependencies. All fine tuning of rules and algorithms was according to the UD training data. Results only pertain to corpora included in UD dev/test datasets.

\section*{Acknowledgments}

\bibliography{custom,references}

\appendix

\section{Example Appendix}
\label{sec:appendix}

This is an appendix.

\end{document}